\documentclass{article}

\usepackage{microtype}
\usepackage{graphicx}
\usepackage{subcaption}
\usepackage{booktabs} 

\usepackage{hyperref}

\usepackage[preprint]{icml2026}

\usepackage{amsmath}
\usepackage{amssymb}
\usepackage{mathtools}
\usepackage{amsthm}

\usepackage{float}
\usepackage{multirow}
\usepackage{tabularx}
\usepackage{caption}
\usepackage{pifont}
\usepackage{placeins}

\usepackage[capitalize,noabbrev]{cleveref}

\theoremstyle{plain}

\theoremstyle{definition}

\theoremstyle{remark}

\usepackage[textsize=tiny]{todonotes}

\graphicspath{
  {2025_winter/introduction/}
  {2025_winter/related_work/}
  {2025_winter/methodology/}
  {2025_winter/experiments/}
  {2025_winter/conclusions/}
  {2025_winter/ablations/}
  {2025_winter/appendix/}
  {2025_winter/figures/}
}

\icmltitlerunning{Semantic Watermarking for Malicious Image Manipulation Detection}

\begin{document}

\twocolumn[
  \icmltitle{Semantic Watermarking for Malicious Image Manipulation Detection}


  \icmlsetsymbol{equal}{*}

  \begin{icmlauthorlist}
    \icmlauthor{Yoonseo Kim}{ku}
    \icmlauthor{Seungwoo Baek}{equal,ku}
    \icmlauthor{Junyoung Park}{equal,ku}
  \end{icmlauthorlist}

  \icmlaffiliation{ku}{Korea University, Seoul, South Korea}

  \icmlcorrespondingauthor{Yoonseo Kim}{seo3167@korea.ac.kr}

  \icmlkeywords{Semantic Watermarking, Image Manipulation Detection, CLIP, Variational Autoencoder, Machine Learning, ICML}

  \vskip 0.3in
]



\printAffiliationsAndNotice{\textsuperscript{*}Equal contribution as co-second authors}

\begin{abstract}
The proliferation of high-fidelity generative editing models has made it
possible to inject violent or sexual content into otherwise ordinary
images while preserving visual plausibility, with concrete consequences
for public discourse and vulnerable populations.
We propose a \textbf{robust semantic watermarking} framework that
reframes the watermark as a recoverable semantic reference rather than
an opaque identifier.
Our framework combines a $\beta$-VAE-based binary watermark
(\textbf{CLIP-VAE}) with explicit \textbf{channel-aware training}---random
bit-flip noise is injected during training so that the decoder learns
graceful degradation under the noisy watermarking channel.
As a downstream application, a lightweight module \textbf{SDA-Net}
uses the recovered semantic embedding to expose not only whether but
\emph{in which} semantic direction an image has been altered.
In a 5-way comparison against representative binary hashing baselines
(SimHash, ITQ, HashNet, and their robust-MLP variants), CLIP-VAE
achieves the highest reconstruction cosine similarity to the original
CLIP embedding under realistic InstructPix2Pix bit-error rates, and
uniquely supports direction-of-drift detection---a forensic complement
to existing content-moderation pipelines.
\end{abstract}

\section{Introduction}

\begin{figure}[t]
    \centering
    \includegraphics[width=0.8\columnwidth]{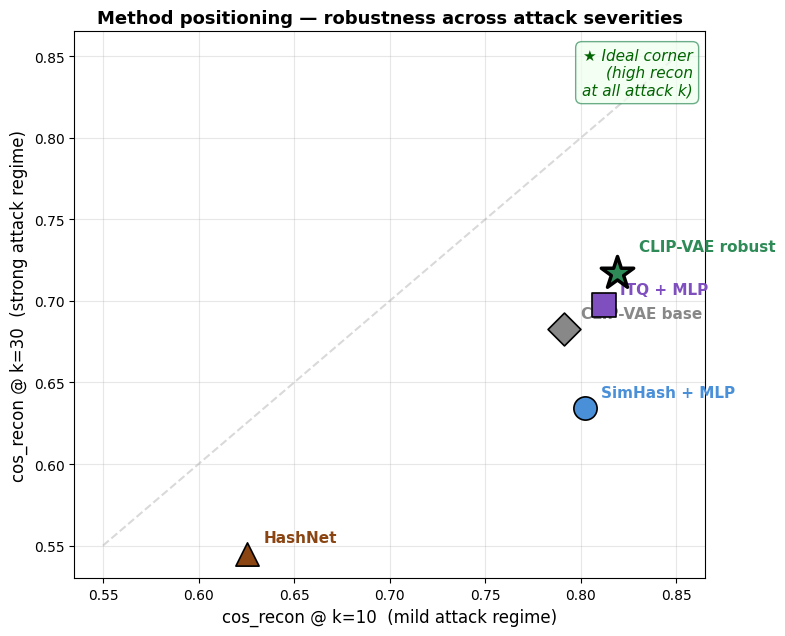}
    \caption{\textbf{Method positioning across attack severities.}
    Five binary semantic-encoding methods (SimHash, ITQ, HashNet,
    CLIP-VAE base, and our channel-aware CLIP-VAE) are plotted in the
    (cos$\,@\,k\!=\!10$, cos$\,@\,k\!=\!30$) plane, summarizing CLIP
    reconstruction quality under mild and strong channel noise.
    The top-right corner (star marker) is ideal: high reconstruction
    at every attack severity.
    Our channel-aware CLIP-VAE sits closest to this corner,
    outperforming all baselines in the realistic InstructPix2Pix BER
    regime; full setup and analysis are in \cref{sec:bitflip}.}
    \label{fig:positioning}
\end{figure}

The recent surge of high-fidelity generative
models~\cite{rombach2022stablediffusion, ramesh2022dalle2, podell2023sdxl, saharia2022imagen}
and image-to-image editing
models~\cite{brooks2023instructpix2pix, meng2022sdedit, hertz2022prompt2prompt, kawar2023imagic}
has made it possible to inject violent or sexual content into otherwise
ordinary scenes while preserving visual plausibility.
The societal cost is no longer hypothetical:
in May 2023, a fabricated AI-generated image of a Pentagon explosion
spread across social media within minutes and briefly moved U.S.\
equity markets.
Because such manipulations evade casual human inspection yet carry
concrete consequences for public discourse and institutional trust,
methods that can reliably analyze \emph{semantic} changes in images
have become an urgent component of trustworthy AI infrastructure.

Once an image has been manipulated, the original semantic content is
structurally lost from the pixel evidence: only the post-edit image
remains observable, and no purely passive analysis can recover what
the image \emph{was}.
A natural first instinct is therefore to deploy classifiers that flag
manipulated content directly from pixels.
Such classifiers, however, operate only on the \emph{current} image
and have no access to its pre-edit state: stylistic edits that
gradually push a benign scene toward violent or sexual content can
register as benign on a frame-by-frame basis, and adversarially crafted
perturbations can evade detection at inference time.
This motivates a complementary approach that anchors a \emph{source-time}
semantic reference into the image itself, against which any later edit
can be compared.
Crucially, this pre-edit information cannot be reconstructed from the
manipulated image alone---it must be deliberately encoded at the moment
of capture or distribution, making a watermark the structurally unique
carrier of the image's original semantic state.

Digital watermarking, then, is a natural defense, allowing information
about the original image to persist after editing.
However, the dominant focus of robust watermarking has been on
\emph{survivability} of the watermark under aggressive edits, rather
than on the \emph{semantic content} of the payload.
A representative line of work---HiDDeN~\cite{zhu2018hidden},
StegaStamp~\cite{tancik2020stegastamp}, RoSteALS~\cite{bui2023rosteals},
Stable Signature~\cite{fernandez2023stablesignature},
Tree-Ring Watermarks~\cite{wen2023treering},
Gaussian Shading~\cite{yang2024gaussianshading}, and
VINE~\cite{lu2024robust}---has substantially advanced robustness, yet
treats the watermark payload as an \emph{opaque identifier}:
a random bit string for ownership verification.
Concurrent work such as SEAL~\cite{arabi2025seal},
SemaMark~\cite{ren2024semamark}, and SWIFT~\cite{evennou2024swift}
(which embeds image captions as real-valued vectors) explores
semantic payloads, but each makes design choices that diverge from a
binary, invertible, channel-robust formulation.
This leaves content moderators able to verify \emph{whether} an image
is authentic, but unable to answer the forensic question of \emph{what}
changed semantically---a gap that becomes critical when the manipulated
content is targeted violence or sexual material.

A natural baseline for binary semantic payloads is the rich literature
of \emph{semantic hashing}, which compresses high-dimensional
embeddings into compact binary codes.
This literature spans three design philosophies that we adopt as
baselines:
(i) \textbf{random projection} (SimHash~\cite{charikar2002simhash},
Spectral Hashing~\cite{weiss2008spectralhashing}),
(ii) \textbf{linear learned hashing} (Iterative
Quantization~\cite{gong2011itq}),
and (iii) \textbf{deep learned hashing}
(HashNet~\cite{cao2017hashnet}, DSH~\cite{liu2016dsh},
VDSH~\cite{chua2018vdsh}).
However, all of these methods are designed for retrieval and are
\emph{non-invertible}---the binary code supports only
Hamming-distance comparison and cannot reconstruct a semantic
embedding for downstream analysis.
Recent neural compressors such as LLMZip~\cite{valmeekam2023llmzip}
go further but produce variable-length codes that collapse
catastrophically under single-bit errors, making them incompatible
with the noisy fixed-capacity channel of practical watermarking.

We take a different perspective and treat the watermark as a
recoverable \emph{semantic reference} rather than an opaque identifier.
Our approach builds on three observations.
\emph{First}, CLIP image embeddings~\cite{radford2021clip} provide a
perceptually-aligned semantic representation~\cite{hessel2021clipscore}
that retains enough information to drive image
reconstruction~\cite{ramesh2022dalle2}.
\emph{Second}, because CLIP embeddings are L2-normalized, semantic
similarity is captured by their angular
structure~\cite{wang2020alignment}, which sign-based binarization of
zero-centered latents preserves---a property formalized by random
hyperplane rounding~\cite{charikar2002simhash, goemans1995maxcut}.
\emph{Third}, a $\beta$-VAE~\cite{kingma2014vae, higgins2017betavae}
regularizes the latent toward an isotropic Gaussian prior, producing
the zero-centered, well-distributed latents that sign binarization
requires and---crucially---enabling \emph{invertible} reconstruction
from the binary code through latent statistics rescaling.
A plain autoencoder lacks this regularization, and existing deep
hashing methods~\cite{cao2017hashnet, liu2016dsh, chua2018vdsh} reuse
similar architectures only for retrieval, not for invertible recovery.

Building on these observations and on the VINE~\cite{lu2024robust}
robust watermarking backbone, our central contribution is
\textbf{CLIP-VAE}, a robust binary semantic watermark.
To demonstrate the downstream utility of the recovered semantic
embedding, we additionally present \textbf{SDA-Net}, a lightweight
prototype-based application module for direction-of-drift detection.
We further introduce \textbf{channel-aware training}: random bit flips
are injected into the binarized latent during VAE training, forcing
the decoder to learn graceful degradation under channel noise.
Unlike pixel-level perturbation training in
HiDDeN~\cite{zhu2018hidden} and StegaStamp~\cite{tancik2020stegastamp},
which models robustness in image space, our channel-aware training
operates \emph{post-binarization} in the code space, directly modeling
the noisy fixed-capacity channel that real watermark decoders produce---a
mechanism unavailable to random-projection methods like SimHash and
not exploited by existing deep hashing baselines.
In a 5-way comparison with SimHash, ITQ, HashNet, and their
robust-MLP variants, CLIP-VAE achieves the highest reconstruction
cosine similarity to the original CLIP embedding under realistic
bit-error rates corresponding to InstructPix2Pix
attacks (\(k\!=\!5\)--\(30\) flipped bits out of 100).
Beyond reconstruction quality, our framework uniquely supports
\emph{direction-of-drift detection} via SDA-Net---identifying not only
whether but in which semantic category an image has been altered, a
capability not provided by any binary hashing baseline.

\section{Related Work}
\label{sec:related_work}

\paragraph{Robust Image Watermarking.}
The challenge of preserving watermarks under aggressive image editing has
driven a steady evolution of robust watermarking methods.
HiDDeN~\cite{zhu2018hidden} introduced an encoder--decoder framework that
learns invisible perturbations resilient to common distortions, while
StegaStamp~\cite{tancik2020stegastamp} extended this paradigm to physical
capture by training against a wide range of imaging artifacts.
RoSteALS~\cite{bui2023rosteals} embedded watermarks via the latent space
of pretrained autoencoders for improved imperceptibility.
More recent methods target the rapidly improving generative editing
landscape: Stable Signature~\cite{fernandez2023stablesignature} fine-tunes
the decoder of latent diffusion models so that all generated images carry
a verifiable signature, Tree-Ring Watermarks~\cite{wen2023treering}
embed signals directly into the initial Gaussian noise of diffusion
models, and Gaussian Shading~\cite{yang2024gaussianshading} achieves
performance-lossless watermarking by aligning the watermark with the
model's noise distribution.
VINE~\cite{lu2024robust}, on which our work builds, demonstrates
100-bit binary watermark survivability under both local and global
generative edits.
Across this rapid progress, payloads remain \emph{random identifiers}---
robustness is the metric of success, semantic content of the payload is
not considered.

\paragraph{Concurrent Semantic-Payload Watermarking.}
SWIFT~\cite{evennou2024swift} is the closest concurrent work to ours
and similarly reframes the watermark payload as a semantic carrier.
The two methods differ in design choices rather than in fundamental
philosophy: SWIFT encodes image captions as high-dimensional
real-valued vectors via a modified HiDDeN backbone, while we encode
CLIP image embeddings as 100-bit binary watermarks via VINE.
Our binary representation is natively compatible with the
high-robustness binary channels of modern robust watermarking, and our
framework introduces channel-aware learning that explicitly optimizes
for bit-flip noise---a direction not explored in SWIFT.
SWIFT also targets general image authentication, whereas our framework
is explicitly oriented toward detecting malicious semantic categories
through prototype-based drift analysis, aligning it directly with
content-moderation pipelines.
We do not provide a direct empirical comparison since SWIFT's
caption-based pipeline requires a separate captioner and HiDDeN
backbone that differ from our setup; constructing a unified benchmark
is left for future work.

\FloatBarrier
\begin{figure*}[t]
    \centering
    \includegraphics[width=\textwidth]{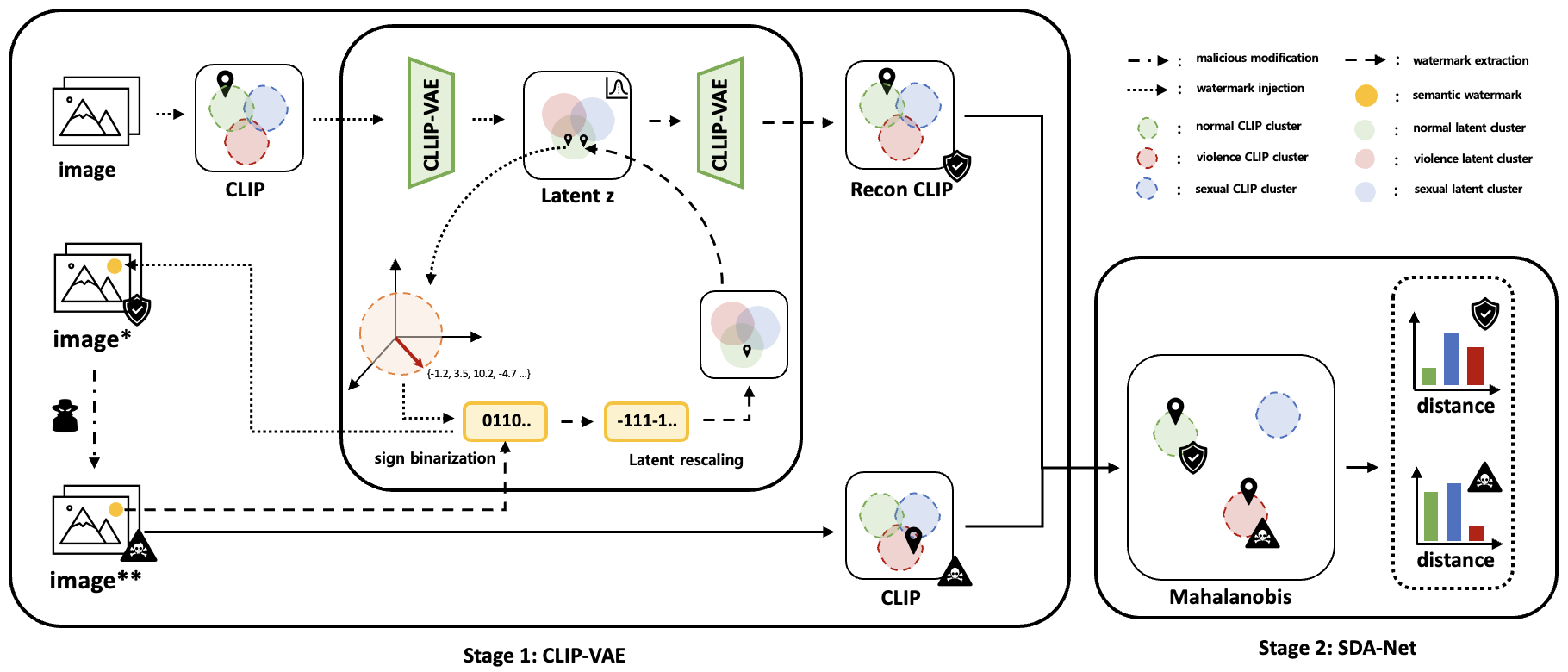}
    \caption{
    CLIP-VAE--based semantic watermarking and drift detection.
    Semantic information is embedded into CLIP-VAE latent representations via sign-based binarization and reconstructed under malicious image manipulation through distribution-aware latent rescaling.
    Semantic drift is detected by comparing reconstructed and manipulated CLIP embeddings in a class-conditional semantic space.
    }
    \label{fig:overall}
\end{figure*}

\paragraph{Semantic Hashing and Neural Compression.}
A separate line of work compresses semantic content into compact binary
codes for efficient retrieval.
SimHash~\cite{charikar2002simhash} produces fingerprints via random
hyperplane projections, and later locality-sensitive variants such as
Spectral Hashing~\cite{weiss2008spectralhashing} and Iterative
Quantization~\cite{gong2011itq} introduce learned projections.
Deep hashing methods including DSH~\cite{liu2016dsh},
HashNet~\cite{cao2017hashnet}, and VDSH~\cite{chua2018vdsh} use learned
representations to produce binary codes whose Hamming distances reflect
semantic similarity, with VDSH in particular employing variational inference.
These methods, however, are \emph{non-invertible}: the binary code
supports only Hamming-distance comparison and cannot be decoded back to
a semantic representation.
Recent neural compressors such as LLMZip~\cite{valmeekam2023llmzip}
achieve near-entropy-rate codes by combining large language models with
arithmetic coding, but produce variable-length outputs that decode
catastrophically under bit errors---incompatible with watermarking's
noisy fixed-capacity channel.
Our framework is the first, to our knowledge, to combine the binary
structure of semantic hashing with an \emph{invertible} mapping back to
semantic embeddings, adapted specifically for the watermarking channel.

\paragraph{Image Manipulation Detection.}
Detecting altered or synthetic images has long been studied as a passive
forensic problem, with early work focused on detecting GAN-generated
faces and deepfakes through frequency-domain artifacts and more recent
methods targeting diffusion-generated content via pixel-space classifiers.
Such passive approaches face a structural limitation: as generative
models improve, the visual gap between real and synthetic narrows.
Active forensic methods---those that embed information into the source
image to enable later verification---sidestep this arms race by
anchoring authenticity at the moment of capture or distribution.
Our work belongs to the active family but goes beyond binary
authentication: rather than only detecting whether an image was tampered
with, we recover a semantic reference of the original image and quantify
the \emph{direction} of any subsequent semantic shift, providing
finer-grained signals useful for content moderation rather than just
content provenance.

\paragraph{Trustworthy AI and Content Moderation.}
Content moderation systems for harmful imagery typically rely on
classifiers trained directly on the moderated content, an approach that
struggles with adversarially edited images that shift only subtle
semantic cues.
Recent work on trustworthy AI emphasizes the importance of complementary
forensic signals that allow moderators to verify the original semantic
intent of an image, particularly in pipelines that handle sensitive
categories such as violence or sexual content.
We position our framework as a forensic complement to such moderation
systems: by preserving a recoverable semantic anchor through an invisible
watermark and exposing the direction of any post-hoc drift, it enables
moderators to flag manipulations that would otherwise pass classifier-only
checks.
This places semantic watermarking within a broader trustworthy-AI
infrastructure rather than treating it as a standalone authentication tool.

\section{Methodology}
Our framework consists of two components (Figure~\ref{fig:overall}): \textbf{CLIP-VAE}, our central contribution, which encodes a CLIP embedding into a 100-bit binary watermark robust to channel noise; and \textbf{SDA-Net}, a lightweight application module that uses the recovered embedding for prototype-based direction-of-drift detection.

\subsection{CLIP-VAE}
We propose \textbf{CLIP-VAE}, a variational autoencoder operating in the CLIP embedding space
that compresses an image's semantics into a compact \textbf{100-bit binary watermark}.

Given an image, we first extract a CLIP image embedding
$\mathbf{x} \in \mathbb{R}^{512}$ using a pretrained CLIP encoder,
which is L2-normalized following the standard CLIP embedding geometry.
The encoder maps $\mathbf{x}$ into a latent representation
$z \in \mathbb{R}^{100}$ via variational inference.
The decoder then maps the latent vector $z$ to a reconstructed embedding
$f_{\text{dec}}(z) \in \mathbb{R}^{512}$.
To ensure consistency with the angular geometry of CLIP embeddings,
we apply L2 normalization \emph{after} decoding, yielding
$\hat{\mathbf{x}} = f_{\text{dec}}(z) / \| f_{\text{dec}}(z) \|_2$.
The resulting $\hat{\mathbf{x}}$ serves as a semantic anchor,
which is compared against the CLIP embedding of a potentially manipulated image
to detect semantic drift.

\paragraph{Training objective.}
We optimize a compact $\beta$-VAE objective that prioritizes semantic fidelity in CLIP space, with the KL regularization weight set to $\beta = 0.01$.
\begin{equation}
\mathcal{L}_{\text{CLIP-VAE}}
= \underbrace{1-\cos(\hat{\mathbf{x}},\mathbf{x})}_{\mathcal{L}_{\text{recon}}}
\;+\; \beta\,\mathcal{L}_{\text{KL}},
\label{eq:clipvae_loss}
\end{equation}

\paragraph{Latent to Bits.}
Our key step is to turn the continuous latent $z$ into a robust \textbf{bit-level semantic watermark}
$b\in\{0,1\}^{D}$ ($D{=}100$) by preserving only the sign pattern:
\begin{equation}
b_i =
\begin{cases}
1 & \text{if } z_i > 0,\\
0 & \text{otherwise}.
\end{cases}
\label{eq:latent_to_bits}
\end{equation}
This binarization discards magnitude but retains directional information, which is well-aligned with the geometry of CLIP embeddings (semantic similarity is largely captured by angles).

\paragraph{Bits to Latent.}
From an extracted watermark $\hat{b}$, we reconstruct a latent proxy $\hat{z}$ in two steps.
First, we map bits back to signs:
\begin{equation}
s_i = 2\hat{b}_i - 1,\quad s\in\{-1,+1\}^{D}
\label{eq:bits_to_sign}
\end{equation}
Then, we restore scale using \textbf{learned latent statistics} (computed once from training latents):
\begin{equation}
\hat{z} = \mu_{\text{latent}} + s \odot \sigma_{\text{latent}}
\label{eq:sign_to_latent}
\end{equation}
where $\mu_{\text{latent}}, \sigma_{\text{latent}} \in \mathbb{R}^{D}$ are the empirical mean and standard deviation of the latent distribution.
This makes reconstruction stable: even when some bits flip, the recovered $\hat{z}$ remains a plausible latent sample, enabling reliable decoding to $\hat{\mathbf{x}}$ and downstream semantic drift comparison.

\paragraph{Channel-Aware Training.}
A standard $\beta$-VAE is trained only on \emph{clean} latents, leaving
the decoder unaware of the bit-flip noise that the watermarking channel
inevitably produces.
We address this with \emph{channel-aware training}: at each training
step, we sample $k\!\sim\!\mathcal{U}(0, k_{\max})$ bits and randomly
flip them in the binarized latent $b$ before reconstructing
$\hat{z} = \mu_\text{latent} + s_\text{noisy} \odot \sigma_\text{latent}$,
then minimize $1-\cos(\hat{\mathbf{x}}, \mathbf{x})$ against the
\emph{original} CLIP embedding.
A straight-through estimator for the sign function preserves gradient
flow through binarization.
We use a curriculum schedule with $k_{\max}\!=\!10$, which our ablation
(\cref{sec:appendix_component_ablation,sec:appendix_hp_sweep}) confirms
improves bit-flip robustness without sacrificing clean reconstruction.
This learning-based channel awareness is fundamentally unavailable to
random-projection methods such as
SimHash~\cite{charikar2002simhash}.

\subsection{Semantic Drift-Aligned Network}
We propose the \textbf{Semantic Drift-Aligned Network (SDA-Net)} to quantify
semantic drift between the preserved watermark semantics and the manipulated
image content.
Instead of producing discrete classification labels, SDA-Net learns a
class-conditional latent geometry in which the \emph{direction} of semantic
change can be measured against learned class prototypes,
exposing not only whether but in which semantic category an image has shifted.

\paragraph{Architecture and Prototypes.}
SDA-Net is a three-layer fully-connected variational encoder
$\mathbb{R}^{512}\!\to\!\mathbb{R}^{64}$
(BatchNorm, ReLU, Dropout(0.3) at each stage) producing
$(\mu,\log\sigma^2)$, with $z\!=\!\mu+\sigma\odot\epsilon$.
Each semantic class
$c\!\in\!\{\text{Normal},\text{Violence},\text{Sexual}\}$ is a
learnable prototype $(\mu_c,\Sigma_c)$ with diagonal
$\Sigma_c\!=\!\mathrm{diag}(\exp(\log\sigma_c^2))$.
The Mahalanobis distance
$d^2(z,c)\!=\!\sum_i (z_i-\mu_{c,i})^2 / \sigma_{c,i}^2$ feeds a
temperature-scaled softmax yielding class scores that vary smoothly
with latent proximity to each prototype.
This proximity reflects \emph{class affiliation}---how typical the
latent is of a category---rather than within-class intensity, a
distinction we revisit in \cref{sec:limitations}.

\paragraph{Training and Drift Measurement.}
SDA-Net is trained with a joint objective combining cross-entropy
(label smoothing 0.1), KL regularization toward $\mathcal{N}(0,I)$,
and a supervised contrastive loss
($\lambda_\text{cls}\!=\!1.0$, $\lambda_\text{KL}\!=\!0.01$,
$\lambda_\text{SCL}\!=\!0.5$, $\tau\!=\!0.07$);
class prototypes are updated via EMA ($m\!=\!0.9$).
Full training details are in \cref{sec:appendix_sdanet_train}.
At inference, given reconstructed and manipulated embeddings
$z_\text{recon}$ and $z_\text{manip}$, we report
\emph{drift magnitude}
$\Delta_{\text{latent}}\!=\!\|z_{\text{recon}}-z_{\text{manip}}\|_2$
and \emph{per-class distance change}
$\Delta_c\!=\!d(z_{\text{manip}},\mu_c)-d(z_{\text{recon}},\mu_c)$;
a negative $\Delta_c$ indicates motion toward class $c$, often
revealing manipulation before the predicted class crosses a category
boundary.

\section{Experiments}
We conduct experiments to evaluate the proposed framework from three complementary perspectives:
(1) semantic fidelity of reconstructed embeddings on held-out test data,
(2) robustness of semantic preservation across different content categories,
and (3) effectiveness of semantic manipulation detection under adversarial image editing.

Our experiments are conducted on a dataset assembled from publicly
released benchmarks for harmful-content research,
in keeping with the ethical handling of sensitive imagery.
To reduce semantic ambiguity arising from unclear category boundaries,
we focus on two malicious content categories:
\textit{Violence} and \textit{Sexual}.
The final dataset consists of 8,000 images,
including 4,000 normal images from a publicly available news image
collection~\cite{news_dataset},
2,000 violent images from the HOD~\cite{ha2023hod} and
T2VS~\cite{yeh2024t2vs} benchmarks,
and 2,000 sexual images from a publicly distributed adult-content
dataset~\cite{adult_dataset}.
The dataset is split into training and test sets with an 8:2 ratio.
We do not collect, scrape, or redistribute any additional sensitive
content beyond what is already released by these benchmarks under their
respective licenses; no human annotation of new sensitive content was
performed; and our reproducibility release contains only the
preprocessing pipeline and references to the original public datasets,
not the images themselves.

Throughout all experiments, we rely on CLIP image embeddings
and cosine similarity as the primary semantic metric,
as CLIP predominantly encodes semantic information through embedding direction.

\subsection{CLIP-VAE}

\paragraph{Semantic Preservation.}
We first measure how well the 100-bit watermark preserves the semantic
content of the original CLIP embedding under clean conditions
(no channel noise).
Cosine similarity between original and reconstructed CLIP embeddings
is reported in \cref{tab:clipvae_cosine}; an analogous t-SNE
visualization of the embedding structure is provided in
\cref{sec:appendix_recon}.

\begin{table}[t]
    \centering
    \small
    \caption{Semantic preservation performance on the test set measured by cosine similarity
    between original and reconstructed CLIP embeddings.}
    \label{tab:clipvae_cosine}
    \begin{tabular}{lccc}
        \toprule
        \textbf{Category} & \textbf{Mean} & \textbf{Std.} & \textbf{Range} \\
        \midrule
        Normal   & 0.7948 & 0.0601 & [0.5120, 0.9496] \\
        Violence & 0.8409 & 0.0642 & [0.5678, 0.9552] \\
        Sexual   & 0.8832 & 0.0446 & [0.7165, 0.9540] \\
        \midrule
        \textbf{Overall} & \textbf{0.8285} & \textbf{0.0685} & [0.6943, 0.9626] \\
        \bottomrule
    \end{tabular}
\end{table}

As shown in Table~\ref{tab:clipvae_cosine}, reconstructed CLIP embeddings achieve
a high mean cosine similarity of 0.8285, indicating stable semantic preservation
across categories. Violence and sexual content exhibit higher similarity scores,
suggesting more robust preservation of semantically salient attributes.


\subsection{SDA-Net}

\paragraph{Classification Performance.}
We evaluate SDA-Net on the held-out test set across the three semantic categories.
Table~\ref{tab:sdanet_perclass} reports per-class precision, recall, and F1.
SDA-Net achieves an overall accuracy of \textbf{98.06\%} and an F1 score of
\textbf{0.9802}, with consistently high performance across all classes.
The Sexual class shows the highest precision (0.9949) and the Normal class the
highest recall (0.9875), confirming that the prototype-based formulation
maintains strong discriminative ability while providing the structured latent
space needed for drift analysis.

\begin{table}[t]
\centering
\small
\caption{Per-class classification performance of SDA-Net on the test set.}
\label{tab:sdanet_perclass}
\begin{tabular}{lccc}
\toprule
\textbf{Class} & \textbf{Precision} & \textbf{Recall} & \textbf{F1 Score} \\
\midrule
Normal   & 0.9765 & 0.9875 & 0.9820 \\
Violence & 0.9747 & 0.9626 & 0.9686 \\
Sexual   & 0.9949 & 0.9850 & 0.9899 \\
\midrule
\textbf{Overall} & \textbf{0.9821} & \textbf{0.9784} & \textbf{0.9802} \\
\bottomrule
\end{tabular}
\end{table}

\paragraph{Latent Space Structure.}
Distance distribution analysis confirms that samples from each class exhibit
small distances to their corresponding prototype
(mean intra-class distance: 1.84--2.00) while maintaining substantial
separation from the other prototypes (mean inter-class distance $> 10.5$).
This well-separated geometry provides a reliable coordinate system in which
the direction of semantic drift can be quantified against fixed semantic anchors.


\subsection{Bit-Flip Robustness: 5-Way Comparison}
\label{sec:bitflip}
We compare CLIP-VAE against four binary hashing baselines spanning the
three design philosophies introduced in
\cref{sec:related_work}:
(i) \textbf{SimHash + MLP (robust)}---random Gaussian hyperplanes paired
with a learned MLP decoder trained under bit-flip noise;
(ii) \textbf{ITQ + MLP (robust)}~\cite{gong2011itq}---PCA followed by a
learned orthogonal rotation, also paired with a robust MLP decoder;
(iii) \textbf{HashNet}~\cite{cao2017hashnet}---deep binary
encoder--decoder trained jointly with tanh annealing;
and (iv) \textbf{CLIP-VAE (base)}---our model trained without channel
noise injection.
For each method we obtain a 100-bit code from the test-set CLIP
embedding, randomly flip $k\!\in\!\{0, 5, 10, 20, 30, 50\}$ bits per
sample (10 trials), and measure the cosine similarity between the
original CLIP embedding and its reconstruction from the corrupted bits.
Note that all baselines---including the originally retrieval-oriented
SimHash, ITQ, and HashNet---are equipped with a learned MLP
reconstruction head trained under matched conditions, equalizing the
task to invertible binary encoding; the resulting gap therefore
isolates channel-aware training rather than reflecting any
retrieval-vs-reconstruction task mismatch.

\begin{table}[t]
\centering
\small
\setlength{\tabcolsep}{6pt}
\caption{Bit-flip reconstruction cosine similarity in the realistic
InstructPix2Pix BER regime (mean over 10 random-flip trials,
$\mathrm{std}\!\le\!0.0024$).
Channel-aware CLIP-VAE achieves the highest score across all
practical bit-error rates.
$k\!=\!0$ (clean) and $k\!=\!50$ (near-random) are reported in
\cref{sec:appendix_extremek}.}
\label{tab:bit_flip_5way}
\begin{tabular}{lcccc}
\toprule
$k$ flipped & 5 & 10 & 20 & 30 \\
\midrule
SimHash + MLP (robust)        & 0.824 & 0.802 & 0.734 & 0.634 \\
ITQ + MLP (robust)            & 0.830 & 0.812 & 0.763 & 0.698 \\
HashNet                       & 0.644 & 0.625 & 0.586 & 0.545 \\
CLIP-VAE (base)               & 0.807 & 0.791 & 0.747 & 0.683 \\
\textbf{CLIP-VAE (ours)}      & \textbf{0.830} & \textbf{0.819} & \textbf{0.783} & \textbf{0.717} \\
\bottomrule
\end{tabular}
\end{table}

In the realistic InstructPix2Pix BER range ($k\!=\!5$--$30$), CLIP-VAE
attains the highest cosine similarity at every bit-error rate
(\cref{tab:bit_flip_5way}; full bit-flip curves in
\cref{fig:bit_flip_5way}), with the gap to the strongest baseline
(ITQ + MLP) growing from $0.7\%$ at $k\!=\!10$ to $1.9\%$ at
$k\!=\!30$.
HashNet underperforms uniformly, indicating that deep learned
binarization without explicit channel-aware regularization fails to
convert encoder capacity into channel robustness.
This clear separation in the realistic-BER regime isolates
\emph{channel-aware training} as the source of our advantage,
visualized as the 2D positioning summary in \cref{fig:positioning}
(top of paper).

\subsection{Direction-of-Drift Detection}
\label{sec:drift}
Beyond reconstruction quality, our framework offers a capability not
provided by any binary hashing baseline:
identifying \emph{toward which} semantic category an image has been
manipulated.

\begin{figure}[t]
    \centering
    \includegraphics[width=\columnwidth]{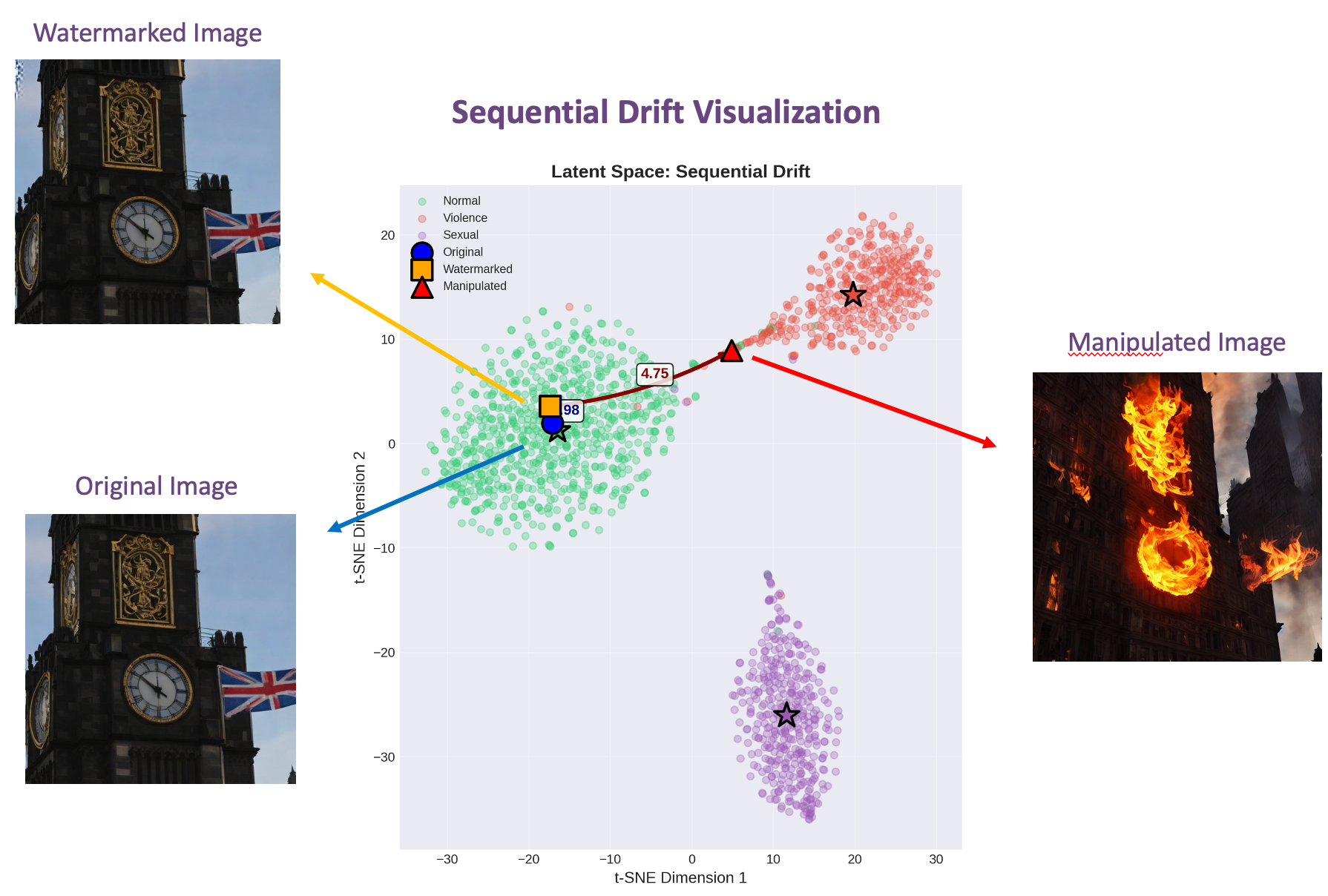}
    \caption{Sequential drift in the SDA-Net latent space.
    The original (blue circle), watermarked (orange square), and
    InstructPix2Pix-edited (red triangle) embeddings are overlaid on
    training-set clusters for Normal (green), Violence (red), and
    Sexual (purple) categories.
    Watermark embedding induces only $\Delta_{\text{latent}}=1.98$,
    while adversarial editing produces $\Delta_{\text{latent}}=4.65$
    directed toward the Violence cluster, exposing the manipulation
    despite visual plausibility.}
    \label{fig:sequential_drift}
\end{figure}

\begin{table}[t]
\centering
\small
\caption{Sequential drift analysis. $\Delta_{\text{latent}}$ denotes
drift magnitude from the original; $\Delta_c$ denotes per-class
distance change (negative = closer to class $c$).
The predicted class remains \textit{Normal} in all three states, but
$\Delta_{\text{Vio}}\!=\!-2.3$ reveals directional drift toward the
Violence prototype before any classifier boundary is crossed.}
\label{tab:sdanet_drift}
\setlength{\tabcolsep}{4pt}
\begin{tabular}{lcccc}
\toprule
\textbf{State} & \textbf{Pred.} & $\boldsymbol{\Delta_{\text{latent}}}$ & $\boldsymbol{\Delta_{\text{Vio}}}$ & $\boldsymbol{\Delta_{\text{Sex}}}$ \\
\midrule
Original    & Normal & ---  & ---    & ---    \\
Watermarked & Normal & 1.98 & $-$0.1 & $+$0.2 \\
Manipulated & Normal & 4.65 & $-$2.3 & $+$0.4 \\
\bottomrule
\end{tabular}
\end{table}

Building on the SDA-Net classification results in
\cref{tab:sdanet_perclass}, we trace an InstructPix2Pix
violence-injection edit through the SDA-Net latent space
(\cref{fig:sequential_drift}, \cref{tab:sdanet_drift}).
The watermarked image remains within the Normal cluster
(\(\Delta_{\text{latent}}=1.98\)), confirming that the watermark itself
does not perturb semantics.
After editing, the image translates by \(\Delta_{\text{latent}}=4.65\)
and the directional signal \(\Delta_{\text{Vio}}=-2.3\) reveals motion
\emph{toward} the Violence prototype.
Crucially, the predicted class remains Normal in all three states---the
directional signal therefore exposes the manipulation \emph{before} the
classifier crosses a category boundary, an early-warning capability
that no binary hashing baseline can produce.

\section{Limitations}
\label{sec:limitations}
Our framework has several limitations that warrant discussion.

\paragraph{Prototype Membership vs.\ Intensity.}
SDA-Net's prototypes represent class centroids rather than within-class
intensity extremes; a drift signal $\Delta_c\!<\!0$ therefore indicates
motion toward the \emph{typical} region of class $c$, not that the
image has become more intensely $c$-like.
A faithful intensity-aware analysis would require ordinal labels
(mild/moderate/severe) that our benchmarks do not provide.

\paragraph{Channel and Modality Limits.}
Our method relies on CLIP, which primarily captures visual semantics,
so manipulations involving textual overlays without visual changes may
go undetected.
The malicious-content scope is also limited to violence and sexual
content; drug-related content or political misinformation is not
considered.
In the realistic InstructPix2Pix attack regime, reconstruction cosine
remains in the $\sim$0.72--0.82 range (\cref{tab:bit_flip_5way}, $k=5$--$30$);
more powerful editing models (e.g., Google Gemini) are expected to
produce higher BER, and a thorough quantification across stronger
editors is left for future work.
Channel-aware training simulates uniform random bit flips approximating
generative-attack noise; transfer to classical signal-domain channels
and stronger robust-watermarking backbones are natural directions.

\paragraph{Linear-Probe Trade-off.}
On linear-probe classification, CLIP-VAE (89\%) is competitive with
but slightly below ITQ + MLP (95\%), which is explicitly optimized for
retrieval-style similarity preservation---a trade-off reflecting our
design priority of \emph{invertible reconstruction under channel noise}.
Adapting flip-noise schedules to class structure is a promising
direction for further gains.

\paragraph{Other Untested Alternatives and Concurrent Work.}
\emph{Hyperspherical VAEs}~\cite{davidson2018hypersphericalvae} and
\emph{Sphere GAN}~\cite{park2019spheregan} offer natural alignment
with CLIP's L2-normalized manifold but use spherical reparameterizations
that are less compatible with sign-based binarization.
\emph{VQ-VAE} provides a natively discrete latent space but suffered
from codebook collapse in our ablation (\cref{sec:ablation}).
\emph{Continuous-valued watermarks} (e.g., SWIFT~\cite{evennou2024swift})
allow higher capacity but sacrifice binary-channel robustness; a unified
benchmark of binary vs.\ real-valued semantic payloads is left for
future work.
The 100-bit payload itself is dictated by the VINE backbone; larger
payloads could carry richer semantic information but require advances
in robust-watermarking capacity beyond the scope of this work.

\section{Ethical Considerations}

\paragraph{Dual-Use Risk.}
By describing how semantic watermarks can detect malicious editing,
we may also inform adversaries who wish to circumvent such detection.
We mitigate this risk in three ways: first, by building on a robust
watermarking backbone (VINE) whose extraction process is open but
whose embedding remains imperceptible; second, by framing semantic
watermarking as a defense-in-depth complement to other moderation
signals rather than a standalone gatekeeper; and third, by encouraging
continued open research on adaptive defenses that can co-evolve with
attack techniques.

\paragraph{False-Positive Risk and Censorship.}
A drift signal that incorrectly flags a normal image as having shifted
toward violence or sexual content could enable inappropriate content
removal or chilling effects on legitimate expression.
For this reason, we treat the directional drift signals
($\Delta_{\text{latent}}$, $\Delta_c$, $\Delta_{\text{score}}$) as
\emph{advisory} rather than authoritative: they are designed to flag
content for human review, and any production deployment should
preserve this human-in-the-loop structure.

\paragraph{Demographic and Cultural Bias.}
Training data for harmful-content detection has historically reflected
the cultural assumptions of its annotators, and the categories we use
(violence, sexual content) carry particularly contested boundaries
across demographic and cultural contexts.
Our framework inherits any biases present in the public datasets we
draw on, and we do not claim that the prototype geometry learned by
SDA-Net is culturally invariant.
Before deployment, downstream practitioners should conduct bias audits
on representative populations and consider category definitions that
reflect the values of the communities the system will serve.

\section{Conclusions}
We presented a robust semantic watermarking framework that reframes the
watermark payload as a recoverable semantic reference rather than an
opaque identifier.
Our framework combines \textbf{CLIP-VAE}---a $\beta$-VAE binary
watermark with sign-based binarization and latent-statistics
rescaling---with \textbf{channel-aware training} that injects bit-flip
noise during training so the decoder learns graceful degradation under
the noisy watermarking channel.
As a downstream application, we further demonstrate that the recovered
semantic embedding supports direction-of-drift detection via
\textbf{SDA-Net}, a lightweight prototype-based module that uniquely
exposes the \emph{direction} of any semantic shift.
In a 5-way comparison with representative binary hashing baselines
(SimHash, ITQ, HashNet, and robust-MLP variants), CLIP-VAE attains the
highest CLIP reconstruction cosine across the realistic InstructPix2Pix
bit-error range.
Future work includes broader malicious-content categories, text-aware
semantic representations, stronger robust-watermarking backbones, and
class-conditional flip-noise schedules.

\section*{Impact Statement}
This work presents a forensic tool for identifying images that have
been semantically manipulated to inject violent or sexual content,
intended as a complement to human-reviewed content-moderation pipelines.
The framework is designed to protect populations particularly affected
by image-based harms---including minors, victims of non-consensual
imagery, and public figures targeted by image manipulation campaigns---
by allowing platforms to verify the original semantic intent of an
image even after editing.
We acknowledge several risks.
Publishing detection techniques may inform adversaries seeking to evade
them; the directional drift signals could be misapplied as authoritative
censorship decisions rather than advisory flags;
and any biases in our training data may propagate into deployment.
We address these risks in detail in our Ethical Considerations section
and emphasize that the framework is intended for human-in-the-loop
moderation, not autonomous content removal.
We encourage continued research on adaptive defenses, fairness audits,
and culturally aware category definitions to ensure that semantic
watermarking serves the trustworthy-AI infrastructure it is meant to
support.

\section*{Reproducibility Statement}
To support reproducibility, we release our complete training and
evaluation code, model configurations, and dataset preprocessing
pipeline at the following anonymous repository:
\begin{center}
\url{https://anonymous.4open.science/r/CLIP-VAE-SDA-Net-25EA}
\end{center}
The repository contains training and evaluation notebooks for both
CLIP-VAE and SDA-Net that reproduce all reported tables and figures,
preprocessing scripts, and detailed instructions for setting up the
environment.
Due to the sensitive nature of the dataset (violence and sexual content),
we do \emph{not} redistribute the images themselves; instead, we provide
preprocessing scripts and references to the original public
datasets~\cite{news_dataset, ha2023hod, yeh2024t2vs, adult_dataset}
so that researchers with appropriate access can reconstruct the exact
dataset used in our experiments.
Hardware details and runtime measurements are included in the repository.

\section*{Use of Generative AI Tools}
In accordance with ICML~2026 policy, we disclose the use of generative
AI tools during the preparation of this manuscript.
We used large language models (LLMs) to assist with manuscript polishing
(grammar correction and English clarity) and to draft portions of the
experimental code, all of which were reviewed, debugged, and validated
by the authors.
All technical content---framework design, experimental methodology,
analyses, results, and conclusions---is the authors' own work, and
all references in this paper were manually verified.

\bibliography{semantic_wm}

@inproceedings{rombach2022stablediffusion,
  title     = {High-Resolution Image Synthesis with Latent Diffusion Models},
  author    = {Rombach, Robin and Blattmann, Andreas and Lorenz, Dominik and Esser, Patrick and Ommer, Bj{\"o}rn},
  booktitle = {Proceedings of the IEEE/CVF Conference on Computer Vision and Pattern Recognition (CVPR)},
  pages     = {10684--10695},
  year      = {2022}
}

@article{ramesh2022dalle2,
  title   = {Hierarchical Text-Conditional Image Generation with {CLIP} Latents},
  author  = {Ramesh, Aditya and Dhariwal, Prafulla and Nichol, Alex and Chu, Casey and Chen, Mark},
  journal = {arXiv preprint arXiv:2204.06125},
  year    = {2022}
}

@inproceedings{podell2023sdxl,
  title     = {{SDXL}: Improving Latent Diffusion Models for High-Resolution Image Synthesis},
  author    = {Podell, Dustin and English, Zion and Lacey, Kyle and Blattmann, Andreas and Dockhorn, Tim and M{\"u}ller, Jonas and Penna, Joe and Rombach, Robin},
  booktitle = {International Conference on Learning Representations (ICLR)},
  year      = {2024}
}

@inproceedings{saharia2022imagen,
  title     = {Photorealistic Text-to-Image Diffusion Models with Deep Language Understanding},
  author    = {Saharia, Chitwan and Chan, William and Saxena, Saurabh and Li, Lala and Whang, Jay and Denton, Emily L. and Ghasemipour, Kamyar and Gontijo Lopes, Raphael and Karagol Ayan, Burcu and Salimans, Tim and Ho, Jonathan and Fleet, David J. and Norouzi, Mohammad},
  booktitle = {Advances in Neural Information Processing Systems (NeurIPS)},
  year      = {2022}
}

@inproceedings{brooks2023instructpix2pix,
  title     = {{InstructPix2Pix}: Learning to Follow Image Editing Instructions},
  author    = {Brooks, Tim and Holynski, Aleksander and Efros, Alexei A.},
  booktitle = {Proceedings of the IEEE/CVF Conference on Computer Vision and Pattern Recognition (CVPR)},
  year      = {2023}
}

@inproceedings{meng2022sdedit,
  title     = {{SDEdit}: Guided Image Synthesis and Editing with Stochastic Differential Equations},
  author    = {Meng, Chenlin and He, Yutong and Song, Yang and Song, Jiaming and Wu, Jiajun and Zhu, Jun-Yan and Ermon, Stefano},
  booktitle = {International Conference on Learning Representations (ICLR)},
  year      = {2022}
}

@inproceedings{hertz2022prompt2prompt,
  title     = {Prompt-to-Prompt Image Editing with Cross-Attention Control},
  author    = {Hertz, Amir and Mokady, Ron and Tenenbaum, Jay and Aberman, Kfir and Pritch, Yael and Cohen-Or, Daniel},
  booktitle = {International Conference on Learning Representations (ICLR)},
  year      = {2023}
}

@inproceedings{kawar2023imagic,
  title     = {{Imagic}: Text-Based Real Image Editing with Diffusion Models},
  author    = {Kawar, Bahjat and Zada, Shiran and Lang, Oran and Tov, Omer and Chang, Huiwen and Dekel, Tali and Mosseri, Inbar and Irani, Michal},
  booktitle = {Proceedings of the IEEE/CVF Conference on Computer Vision and Pattern Recognition (CVPR)},
  year      = {2023}
}

@inproceedings{zhu2018hidden,
  title     = {{HiDDeN}: Hiding Data with Deep Networks},
  author    = {Zhu, Jiren and Kaplan, Russell and Johnson, Justin and Fei-Fei, Li},
  booktitle = {Proceedings of the European Conference on Computer Vision (ECCV)},
  pages     = {657--672},
  year      = {2018}
}

@inproceedings{tancik2020stegastamp,
  title     = {{StegaStamp}: Invisible Hyperlinks in Physical Photographs},
  author    = {Tancik, Matthew and Mildenhall, Ben and Ng, Ren},
  booktitle = {Proceedings of the IEEE/CVF Conference on Computer Vision and Pattern Recognition (CVPR)},
  year      = {2020}
}

@inproceedings{bui2023rosteals,
  title     = {{RoSteALS}: Robust Steganography using Autoencoder Latent Space},
  author    = {Bui, Tu and Agarwal, Shruti and Yu, Ning and Collomosse, John},
  booktitle = {Proceedings of the IEEE/CVF Conference on Computer Vision and Pattern Recognition Workshops (CVPRW)},
  year      = {2023}
}

@inproceedings{fernandez2023stablesignature,
  title     = {The Stable Signature: Rooting Watermarks in Latent Diffusion Models},
  author    = {Fernandez, Pierre and Couairon, Guillaume and J{\'e}gou, Herv{\'e} and Douze, Matthijs and Furon, Teddy},
  booktitle = {Proceedings of the IEEE/CVF International Conference on Computer Vision (ICCV)},
  year      = {2023}
}

@inproceedings{wen2023treering,
  title     = {Tree-Ring Watermarks: Fingerprints for Diffusion Images that are Invisible and Robust},
  author    = {Wen, Yuxin and Kirchenbauer, John and Geiping, Jonas and Goldstein, Tom},
  booktitle = {Advances in Neural Information Processing Systems (NeurIPS)},
  year      = {2023}
}

@inproceedings{yang2024gaussianshading,
  title     = {Gaussian Shading: Provable Performance-Lossless Image Watermarking for Diffusion Models},
  author    = {Yang, Zijin and Zeng, Kai and Chen, Kejiang and Fang, Han and Zhang, Weiming and Yu, Nenghai},
  booktitle = {Proceedings of the IEEE/CVF Conference on Computer Vision and Pattern Recognition (CVPR)},
  year      = {2024}
}

@article{lu2024robust,
  title   = {Robust Watermarking using Generative Priors against Image Editing: From Benchmarking to Advances},
  author  = {Lu, Shilin and Zhou, Zihan and Lu, Jiayou and Zhu, Yuanzhi and Kong, Adams Wai-Kin},
  journal = {arXiv preprint arXiv:2410.18775},
  year    = {2024}
}

@article{arabi2025seal,
  title   = {{SEAL}: Semantic Aware Image Watermarking},
  author  = {Arabi, Kasra and Witter, R. Teal and Hegde, Chinmay and Cohen, Niv},
  journal = {arXiv preprint arXiv:2503.12172},
  year    = {2025}
}

@inproceedings{ren2024semamark,
  title     = {A Robust Semantics-based Watermark for Large Language Models against Paraphrasing},
  author    = {Ren, Jie and Xu, Han and Liu, Yiding and Cui, Yingqian and Wang, Shuaiqiang and Yin, Dawei and Tang, Jiliang},
  booktitle = {Findings of the Association for Computational Linguistics: NAACL},
  year      = {2024}
}

@article{evennou2024swift,
  title   = {{SWIFT}: Semantic Watermarking for Image Forgery Thwarting},
  author  = {Evennou, Gautier and Chappelier, Vivien and Kijak, Ewa and Furon, Teddy},
  journal = {arXiv preprint arXiv:2407.18995},
  year    = {2024}
}

@inproceedings{charikar2002simhash,
  title     = {Similarity Estimation Techniques from Rounding Algorithms},
  author    = {Charikar, Moses S.},
  booktitle = {Proceedings of the Thiry-Fourth Annual ACM Symposium on Theory of Computing (STOC)},
  pages     = {380--388},
  year      = {2002}
}

@inproceedings{weiss2008spectralhashing,
  title     = {Spectral Hashing},
  author    = {Weiss, Yair and Torralba, Antonio and Fergus, Robert},
  booktitle = {Advances in Neural Information Processing Systems (NIPS)},
  year      = {2008}
}

@inproceedings{cao2017hashnet,
  title     = {{HashNet}: Deep Learning to Hash by Continuation},
  author    = {Cao, Zhangjie and Long, Mingsheng and Wang, Jianmin and Yu, Philip S.},
  booktitle = {Proceedings of the IEEE International Conference on Computer Vision (ICCV)},
  pages     = {5608--5617},
  year      = {2017}
}

@inproceedings{gong2011itq,
  title     = {Iterative Quantization: A Procrustean Approach to Learning Binary Codes},
  author    = {Gong, Yunchao and Lazebnik, Svetlana},
  booktitle = {Proceedings of the IEEE Conference on Computer Vision and Pattern Recognition (CVPR)},
  year      = {2011}
}

@inproceedings{liu2016dsh,
  title     = {Deep Supervised Hashing for Fast Image Retrieval},
  author    = {Liu, Haomiao and Wang, Ruiping and Shan, Shiguang and Chen, Xilin},
  booktitle = {Proceedings of the IEEE Conference on Computer Vision and Pattern Recognition (CVPR)},
  pages     = {2064--2072},
  year      = {2016}
}

@inproceedings{chua2018vdsh,
  title     = {Variational Deep Semantic Hashing for Text Documents},
  author    = {Chaidaroon, Suthee and Fang, Yi},
  booktitle = {Proceedings of the 40th International ACM SIGIR Conference on Research and Development in Information Retrieval},
  year      = {2017}
}

@article{valmeekam2023llmzip,
  title   = {{LLMZip}: Lossless Text Compression using Large Language Models},
  author  = {Valmeekam, Chandra Shekhara Kaushik and Narayanan, Krishna and Kalathil, Dileep and Chamberland, Jean-Francois and Shakkottai, Srinivas},
  journal = {arXiv preprint arXiv:2306.04050},
  year    = {2023}
}

@article{goemans1995maxcut,
  title   = {Improved Approximation Algorithms for Maximum Cut and Satisfiability Problems Using Semidefinite Programming},
  author  = {Goemans, Michel X. and Williamson, David P.},
  journal = {Journal of the ACM},
  volume  = {42},
  number  = {6},
  pages   = {1115--1145},
  year    = {1995}
}

@inproceedings{wang2020alignment,
  title     = {Understanding Contrastive Representation Learning through Alignment and Uniformity on the Hypersphere},
  author    = {Wang, Tongzhou and Isola, Phillip},
  booktitle = {International Conference on Machine Learning (ICML)},
  year      = {2020}
}

@inproceedings{radford2021clip,
  title     = {Learning Transferable Visual Models from Natural Language Supervision},
  author    = {Radford, Alec and Kim, Jong Wook and Hallacy, Chris and Ramesh, Aditya and Goh, Gabriel and Agarwal, Sandhini and Sastry, Girish and Askell, Amanda and Mishkin, Pamela and Clark, Jack and Krueger, Gretchen and Sutskever, Ilya},
  booktitle = {Proceedings of the 38th International Conference on Machine Learning (ICML)},
  year      = {2021}
}

@inproceedings{hessel2021clipscore,
  title     = {{CLIPScore}: A Reference-free Evaluation Metric for Image Captioning},
  author    = {Hessel, Jack and Holtzman, Ari and Forbes, Maxwell and Le Bras, Ronan and Choi, Yejin},
  booktitle = {Proceedings of the 2021 Conference on Empirical Methods in Natural Language Processing (EMNLP)},
  year      = {2021}
}

@inproceedings{kingma2014vae,
  title     = {Auto-Encoding Variational {B}ayes},
  author    = {Kingma, Diederik P. and Welling, Max},
  booktitle = {International Conference on Learning Representations (ICLR)},
  year      = {2014}
}

@inproceedings{higgins2017betavae,
  title     = {{$\beta$-VAE}: Learning Basic Visual Concepts with a Constrained Variational Framework},
  author    = {Higgins, Irina and Matthey, Loic and Pal, Arka and Burgess, Christopher and Glorot, Xavier and Botvinick, Matthew and Mohamed, Shakir and Lerchner, Alexander},
  booktitle = {International Conference on Learning Representations (ICLR)},
  year      = {2017}
}

@inproceedings{davidson2018hypersphericalvae,
  title     = {Hyperspherical Variational Auto-Encoders},
  author    = {Davidson, Tim R. and Falorsi, Luca and De Cao, Nicola and Kipf, Thomas and Tomczak, Jakub M.},
  booktitle = {Proceedings of the 34th Conference on Uncertainty in Artificial Intelligence (UAI)},
  year      = {2018}
}

@inproceedings{park2019spheregan,
  title     = {Sphere Generative Adversarial Network Based on Geometric Moment Matching},
  author    = {Park, Sung Woo and Kwon, Junseok},
  booktitle = {Proceedings of the IEEE/CVF Conference on Computer Vision and Pattern Recognition (CVPR)},
  year      = {2019}
}

@misc{news_dataset,
  title        = {News Dataset with Images},
  author       = {{Kaggle}},
  howpublished = {\url{https://www.kaggle.com/datasets/mdkabinhasan/news-dataset-with-images/data}}
}

@inproceedings{ha2023hod,
  title     = {{HOD}: New Harmful Object Detection Benchmarks for Robust Surveillance},
  author    = {Ha, Eungyeom and Kim, Heemook and Hong, Sung Chul and Na, Dongbin},
  booktitle = {Proceedings of the IEEE/CVF Winter Conference on Applications of Computer Vision (WACV) Workshops},
  year      = {2024}
}

@inproceedings{yeh2024t2vs,
  author    = {Chen Yeh and You-Ming Chang and Wei-Chen Chiu and Ning Yu},
  booktitle = {Advances in Neural Information Processing Systems},
  title     = {{T2Vs} Meet {VLMs}: A Scalable Multimodal Dataset for Visual Harmfulness Recognition},
  year      = {2024}
}

@misc{adult_dataset,
  title        = {Adult Content Dataset},
  author       = {{Figshare}},
  howpublished = {\url{https://figshare.com/articles/dataset/Adult_content_dataset/13456484}}
}
\bibliographystyle{icml2026}

\newpage
\appendix
\onecolumn
\section*{Appendices}

\setlength{\textfloatsep}{3pt}
\setlength{\floatsep}{3pt}
\setlength{\intextsep}{3pt}
\setlength{\abovecaptionskip}{4pt}
\setlength{\belowcaptionskip}{2pt}

\section{Ablation Studies}
\label{sec:ablation}
We conduct ablation studies to validate our design choices for CLIP-VAE and SDA-Net.

\subsection{Latent Dimensionality}
\label{subsec:ablation_latent_dim}
Table~\ref{tab:latent_dim_ablation} shows that higher-dimensional latents
yield better CLIP-level reconstruction and category accuracy,
despite a slight decrease in latent-level cosine similarity.
\emph{These results are obtained without watermark quantization},
reflecting the intrinsic semantic capacity of the latent space.
We therefore adopt a 100-dimensional latent configuration.

\begin{table}[H]
\centering
\small
\caption{Effect of latent dimensionality on semantic reconstruction.}
\label{tab:latent_dim_ablation}
\begin{tabularx}{\columnwidth}{c *{3}{>{\centering\arraybackslash}X}}
\toprule
Latent Dim & Latent Cosine $\uparrow$ & CLIP Cosine $\uparrow$ & Category Acc. $\uparrow$ \\
\midrule
50D  & 0.8070 & 0.9525 & 0.8877 \\
100D & 0.8055 & \textbf{0.9648} & \textbf{0.8942} \\
\bottomrule
\end{tabularx}
\end{table}

\subsection{Effect of KL Weight $\beta$}
\label{subsec:ablation_beta}
With $\beta=0.01$, we achieve a well-balanced latent space that preserves
semantic information while maintaining a stable distribution suitable for
binarization.
Smaller values ($\beta = 0$ or $0.001$) leave the latent under-regularized
and unsuitable for sign-based binarization;
larger values ($\beta = 0.1$) over-regularize and destroy the semantic
structure needed for reconstruction.

\begin{table}[H]
\centering
\small
\caption{Effect of KL weight $\beta$ on reconstruction quality and latent regularity.}
\label{tab:beta_ablation}
\begin{tabularx}{\columnwidth}{c *{2}{>{\centering\arraybackslash}X}}
\toprule
$\beta$ & CLIP Cosine Sim $\uparrow$ & KL Divergence $\downarrow$ \\
\midrule
0.0   & 0.8392 & 5.47 \\
0.001 & \textbf{0.8568} & 2.06 \\
0.01  & 0.8339 & 0.70 \\
0.1   & 0.7965 & \textbf{0.19} \\
\bottomrule
\end{tabularx}
\end{table}

\begin{figure}[H]
\centering
\begin{subfigure}{0.49\columnwidth}
    \includegraphics[width=\linewidth]{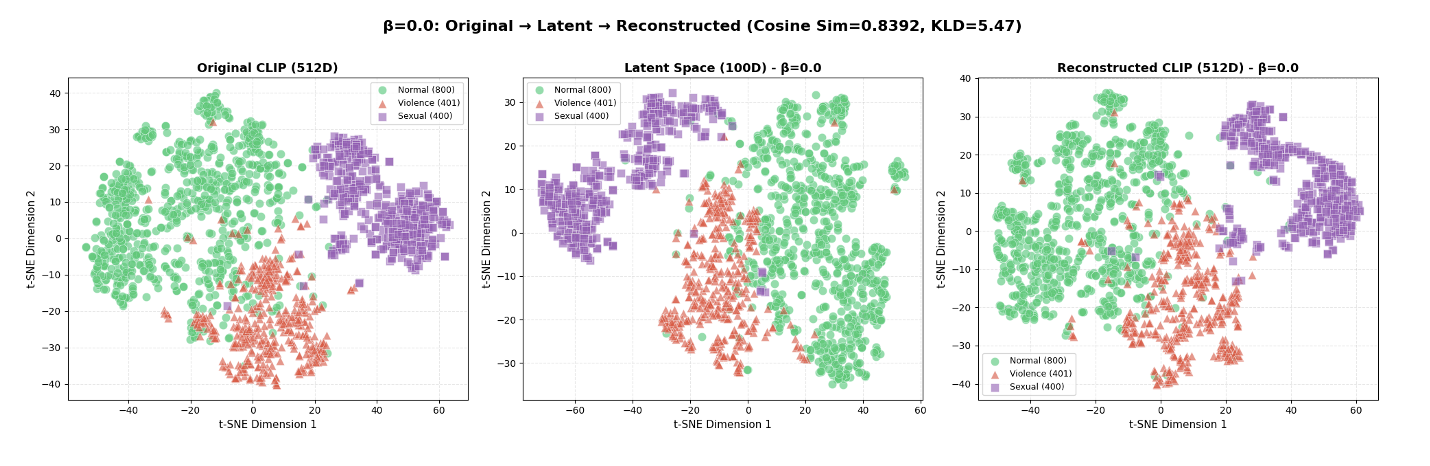}
    \caption{$\beta=0.0$}
    \label{fig:beta_tsne_a}
\end{subfigure}
\hfill
\begin{subfigure}{0.49\columnwidth}
    \includegraphics[width=\linewidth]{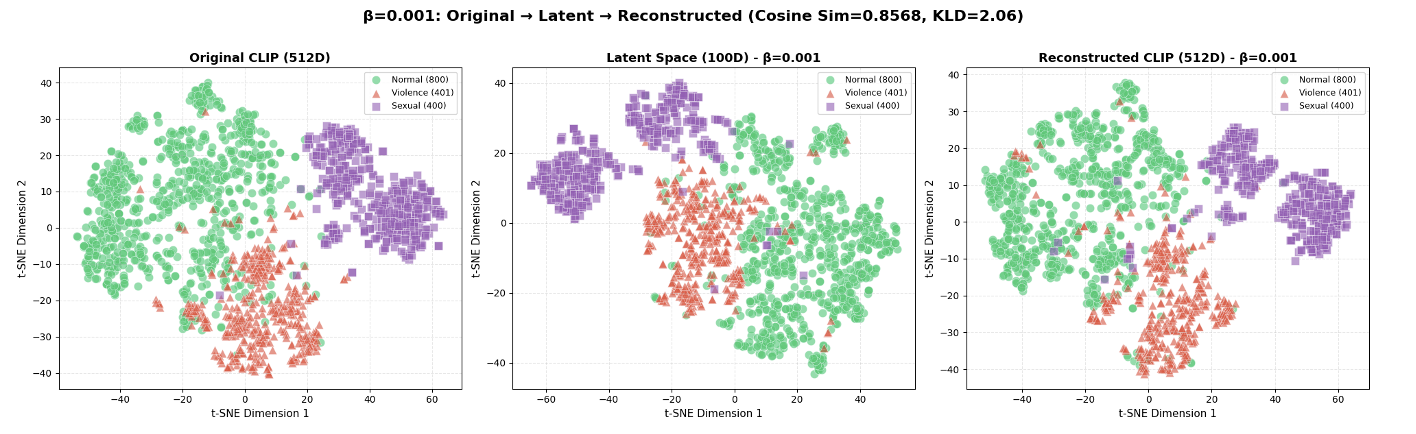}
    \caption{$\beta=0.001$}
    \label{fig:beta_tsne_b}
\end{subfigure}
\\[2pt]
\begin{subfigure}{0.49\columnwidth}
    \includegraphics[width=\linewidth]{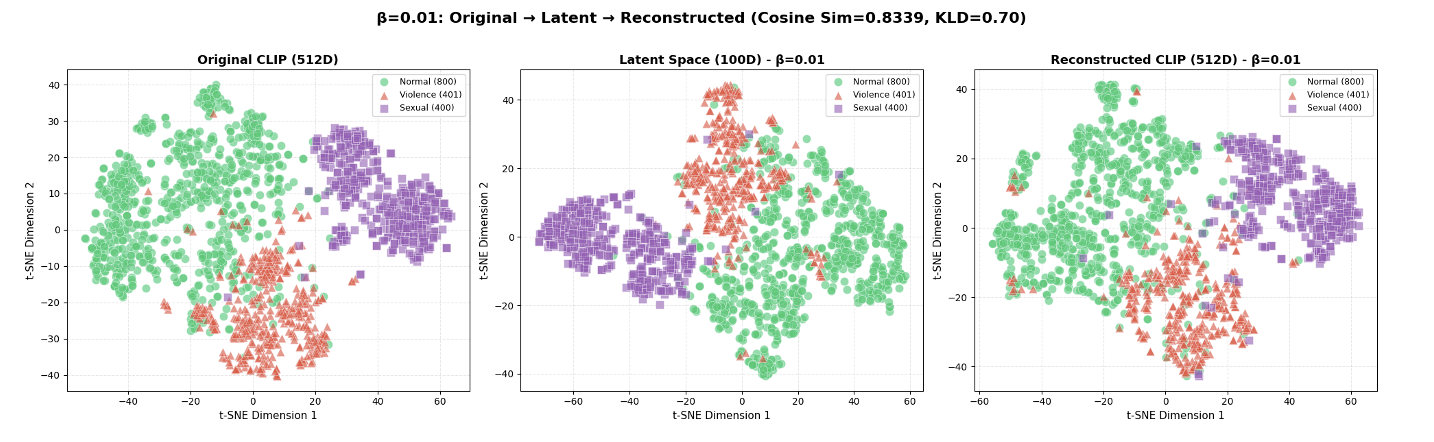}
    \caption{$\beta=0.01$ (chosen)}
    \label{fig:beta_tsne_c}
\end{subfigure}
\hfill
\begin{subfigure}{0.49\columnwidth}
    \includegraphics[width=\linewidth]{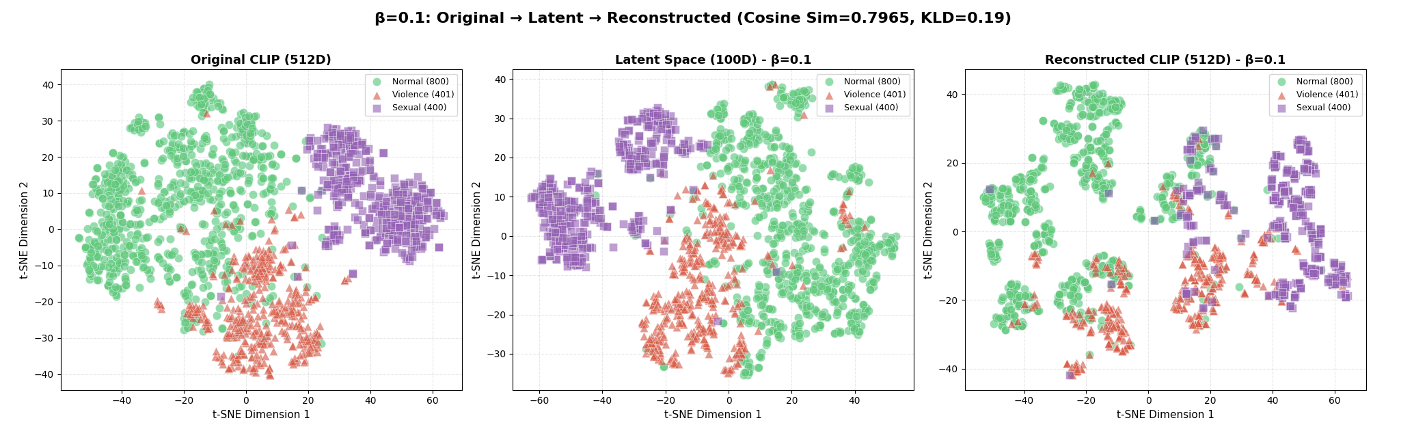}
    \caption{$\beta=0.1$}
    \label{fig:beta_tsne_d}
\end{subfigure}
\caption{Latent space visualizations under different KL weights $\beta$.
With $\beta=0$, the latent is unconstrained and asymmetric;
$\beta=0.001$ leaves the distribution insufficiently regularized for
binarization; $\beta=0.01$ produces the desired well-distributed,
class-separated latent geometry; $\beta=0.1$ over-regularizes and
collapses class structure.}
\label{fig:beta_tsne_grid}
\end{figure}

\subsection{VAE vs.\ VQ-VAE}
\label{subsec:ablation_vqvae}
Given that our watermarking method embeds discrete signals,
we initially hypothesized that the discrete latent structure of VQ-VAE
would be a natural fit.
However, while VQ-VAE enforces clustering around a finite codebook,
its latent representations tend to collapse toward fixed reference points,
limiting fine-grained semantic preservation.
The VAE's continuous latents, regularized by a Gaussian prior,
preserve semantic geometry more faithfully and remain compatible with
sign-based binarization.

\begin{table}[H]
\centering
\small
\caption{Comparison between VAE and VQ-VAE on semantic reconstruction quality.}
\label{tab:vqvae_comparison}
\begin{tabularx}{\columnwidth}{c *{2}{>{\centering\arraybackslash}X}}
\toprule
Model & Cosine Similarity $\uparrow$ & Cosine Distance $\downarrow$ \\
\midrule
VAE    & \textbf{0.8366} & \textbf{0.1634} \\
VQ-VAE & 0.7677 & 0.2323 \\
\bottomrule
\end{tabularx}
\end{table}

\begin{figure}[H]
\centering
\includegraphics[width=0.7\columnwidth]{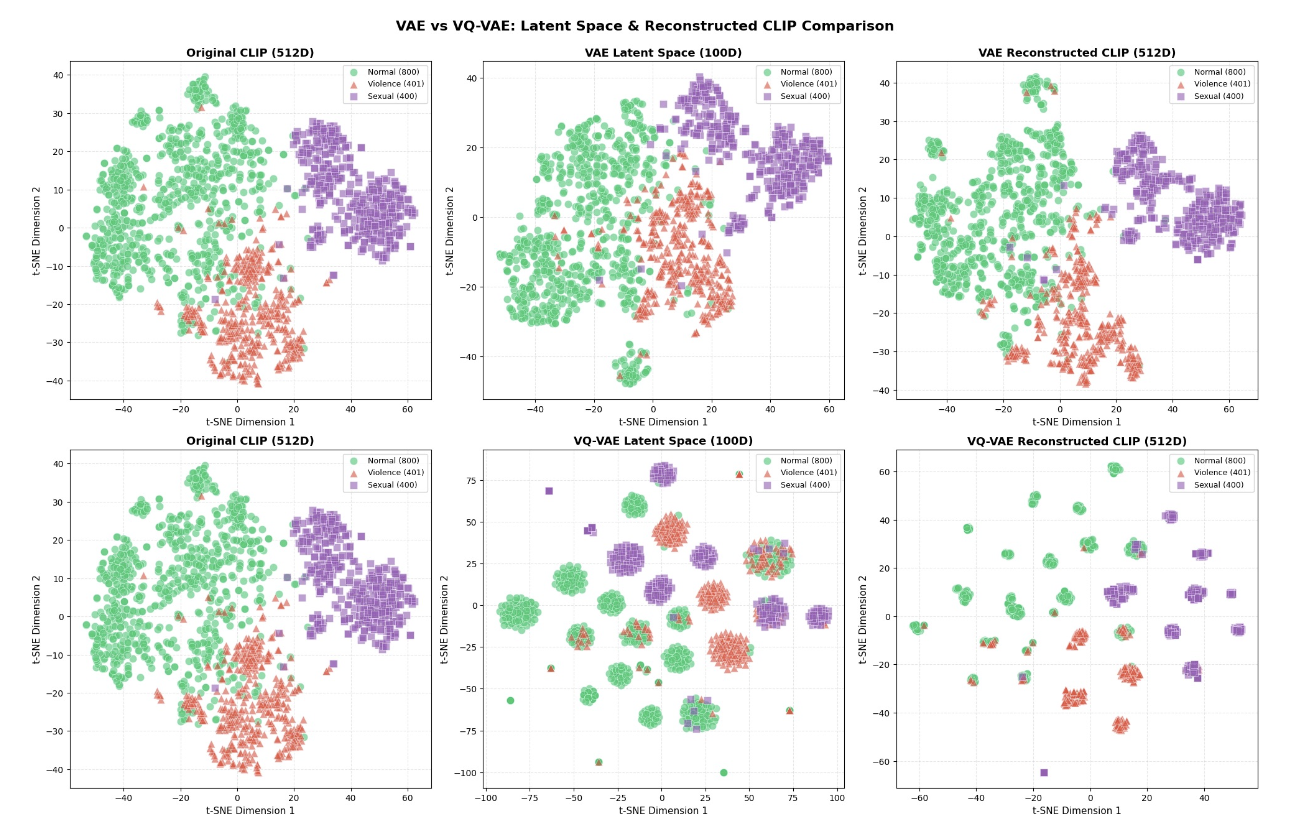}
\caption{Latent space comparison between VAE and VQ-VAE.
VQ-VAE latents cluster tightly around a few codebook entries
(visible codebook collapse), while VAE latents form smooth,
semantically meaningful regions amenable to sign-based binarization.}
\label{fig:vqvae_comparison01}
\end{figure}

\section{Watermark Influence Analysis}
\label{sec:watermark_influence}
We verified that semantic watermarking does not affect CLIP embeddings or image editing outcomes. The CLIP embedding of a watermarked image remains nearly identical to that of the original, and edited images produce indistinguishable embeddings regardless of whether the original was watermarked.

\section{CLIP-VAE Reconstruction Visualization}
\label{sec:appendix_recon}
\begin{figure}[H]
    \centering
    \includegraphics[width=0.7\columnwidth]{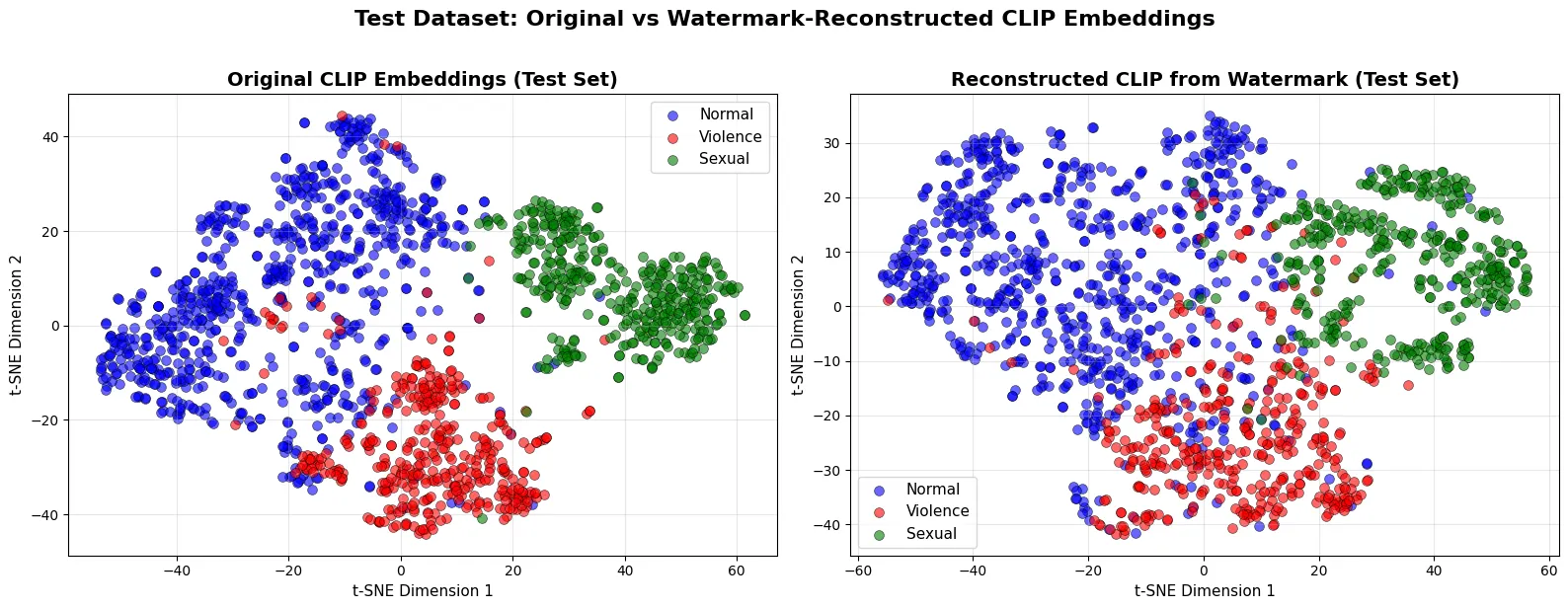}
    \caption{t-SNE visualization of original and reconstructed CLIP
    embeddings on the test set. Reconstructed embeddings largely
    preserve the semantic structure of the originals.}
    \label{fig:clipvae_test_tsne_appendix}
\end{figure}

\Cref{fig:clipvae_test_tsne_appendix} shows a t-SNE visualization of
original and reconstructed CLIP embeddings on the test set.
Reconstructed embeddings remain within their original semantic regions
despite aggressive latent compression into 100 bits, providing a
qualitative complement to the cosine-similarity statistics in
\cref{tab:clipvae_cosine}.

\section{Full Bit-Flip Robustness Curves}
\label{sec:appendix_extremek}
\begin{figure}[H]
    \centering
    \includegraphics[width=0.8\columnwidth]{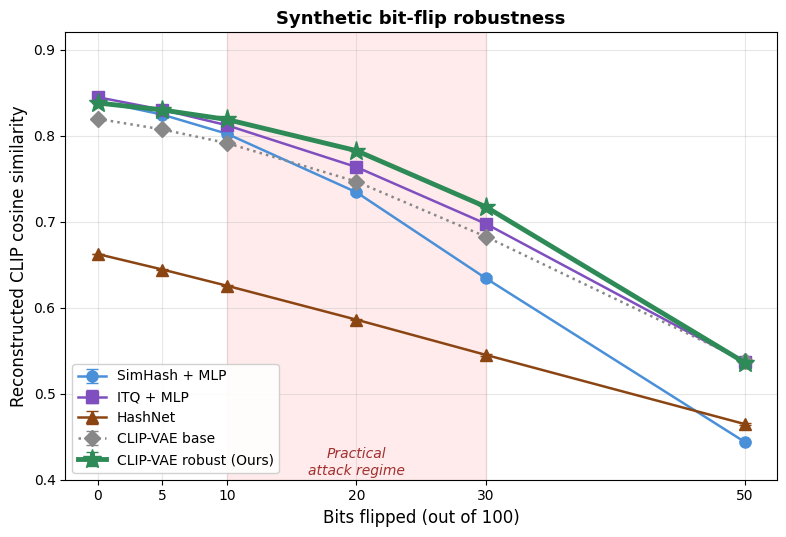}
    \caption{Reconstruction cosine similarity as a function of the
    bit-flip rate $k$ for the five binary hashing methods.
    The shaded region (\(k\!=\!8\)--$28$) corresponds to the BER range
    observed under InstructPix2Pix attacks.
    CLIP-VAE with channel-aware training achieves the highest
    reconstruction in the realistic regime; HashNet underperforms
    uniformly.}
    \label{fig:bit_flip_5way}
\end{figure}

\Cref{fig:bit_flip_5way} shows the full bit-flip robustness curves for
the five methods across $k\!\in\!\{0,5,10,20,30,50\}$.
\Cref{tab:bit_flip_extreme} reports reconstruction cosine at the two
extreme bit-flip rates omitted from the main table.
At $k\!=\!0$ all learned methods cluster near $\sim\!0.84$, confirming
that with a learned decoder even random projections (SimHash + MLP)
can match learned binarization for \emph{clean} reconstruction.
At $k\!=\!50$ all methods converge near $\sim\!0.5$ as reconstruction
is dominated by the latent prior.
Both endpoints are essentially uninformative for distinguishing the
methods; the practical separation occurs in the realistic
$k\!=\!5$--$30$ range reported in \cref{tab:bit_flip_5way}.

\begin{table}[H]
\centering
\small
\setlength{\tabcolsep}{6pt}
\caption{Bit-flip reconstruction cosine at extreme $k$ (clean and
near-random).}
\label{tab:bit_flip_extreme}
\begin{tabular}{lcc}
\toprule
$k$ flipped & 0 (clean) & 50 (near-random) \\
\midrule
SimHash + MLP (robust)        & 0.840 & 0.444 \\
ITQ + MLP (robust)            & 0.845 & 0.537 \\
HashNet                       & 0.662 & 0.465 \\
CLIP-VAE (base)               & 0.819 & 0.538 \\
CLIP-VAE (ours)               & 0.838 & 0.536 \\
\bottomrule
\end{tabular}
\end{table}

\section{SDA-Net Training Hyperparameters}
\label{sec:appendix_sdanet_train}
\Cref{tab:sdanet_hyperparams} summarizes the hyperparameters used to
train SDA-Net.

\begin{table}[H]
\centering
\small
\caption{SDA-Net training hyperparameters.}
\label{tab:sdanet_hyperparams}
\begin{tabular}{lll}
\toprule
\textbf{Component} & \textbf{Setting} & \textbf{Purpose} \\
\midrule
Classification & $\lambda_{\text{cls}}\!=\!1.0$ & CE + label smoothing 0.1 \\
KL Divergence  & $\lambda_{\text{KL}}\!=\!0.01$ & Regularize to $\mathcal{N}(0, I)$ \\
SCL            & $\lambda_{\text{SCL}}\!=\!0.5$ & Class separation ($\tau\!=\!0.07$) \\
Prototype EMA  & $m\!=\!0.9$ & Stabilize prototypes \\
\bottomrule
\end{tabular}
\end{table}

\section{Channel-Aware Training: Component Ablation}
\label{sec:appendix_component_ablation}
We disentangle the two regularizers introduced for channel-aware
training---\emph{flip-noise} (random $k$-bit flip injection at the
binarization stage) and \emph{sign margin} (penalty on near-boundary
$|z_i|<\epsilon$)---by training four variants with the same backbone
and schedule.
\Cref{tab:component_ablation} reports bit-flip robustness
($\cos\!@\!k\!=\!30$) and linear-probe accuracy on the 100-bit code.
Flip-noise alone provides nearly all of the robustness improvement
(\(0.708\) vs.\ \(0.684\) for the no-regularizer baseline) and the
strongest linear-probe accuracy (\(0.94\)).
Adding sign-margin yields a small additional gain in robustness with
a slight cost to linear-probe accuracy.

\begin{table}[H]
\centering
\small
\setlength{\tabcolsep}{8pt}
\caption{Component ablation of channel-aware training.
\emph{flip-noise} alone provides the bulk of the robustness benefit;
\emph{sign-margin} acts as a smaller complementary regularizer.}
\label{tab:component_ablation}
\begin{tabular}{lcc}
\toprule
Variant                              & $\cos\!@\!k\!=\!30$ & LinProbe \\
\midrule
neither                              & 0.684 & 0.854 \\
sign-margin only                     & 0.689 & 0.900 \\
flip-noise only                      & 0.708 & \textbf{0.941} \\
\textbf{flip-noise + sign-margin (ours)} & \textbf{0.702} & 0.918 \\
\bottomrule
\end{tabular}
\end{table}

\begin{figure}[H]
    \centering
    \includegraphics[width=\columnwidth]{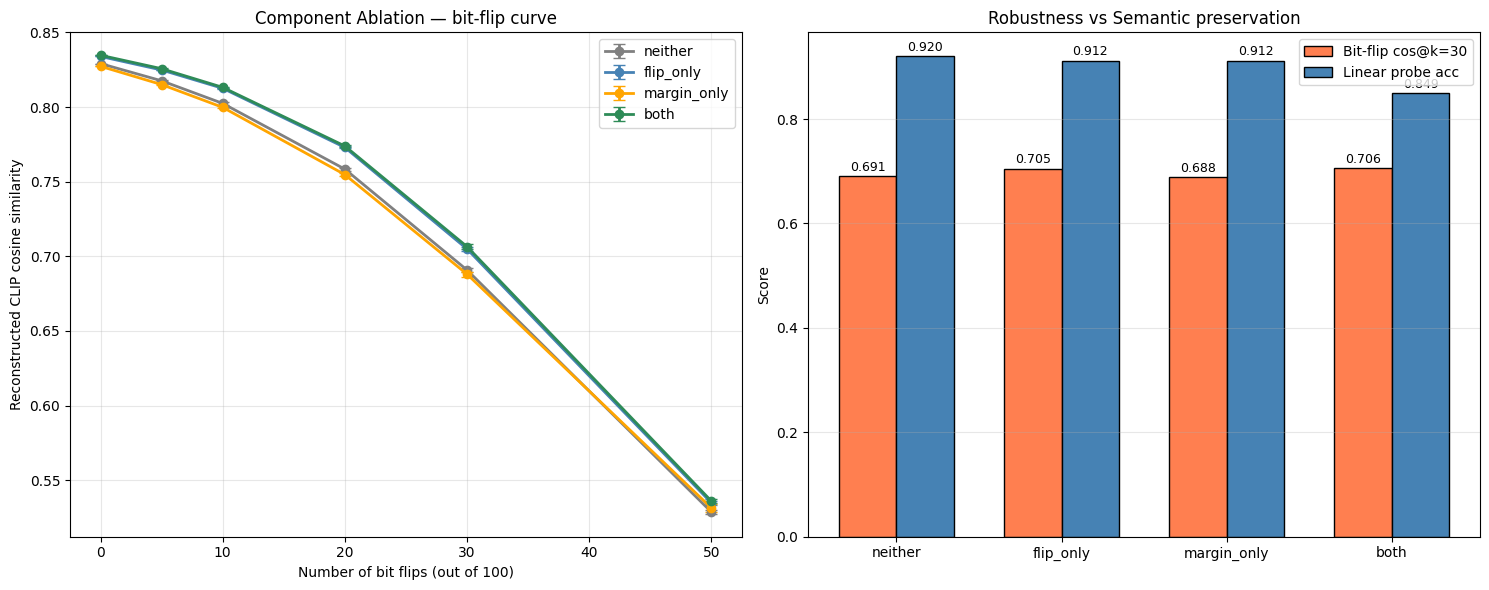}
    \caption{Component ablation visualised as bit-flip robustness curve
    (left) and robustness--semantic preservation summary (right).
    flip-noise injection alone provides the bulk of the bit-flip
    improvement; sign-margin contributes mainly to linear-probe
    accuracy without adding robustness.}
    \label{fig:ablation_curves}
\end{figure}

\begin{figure}[H]
    \centering
    \includegraphics[width=\columnwidth]{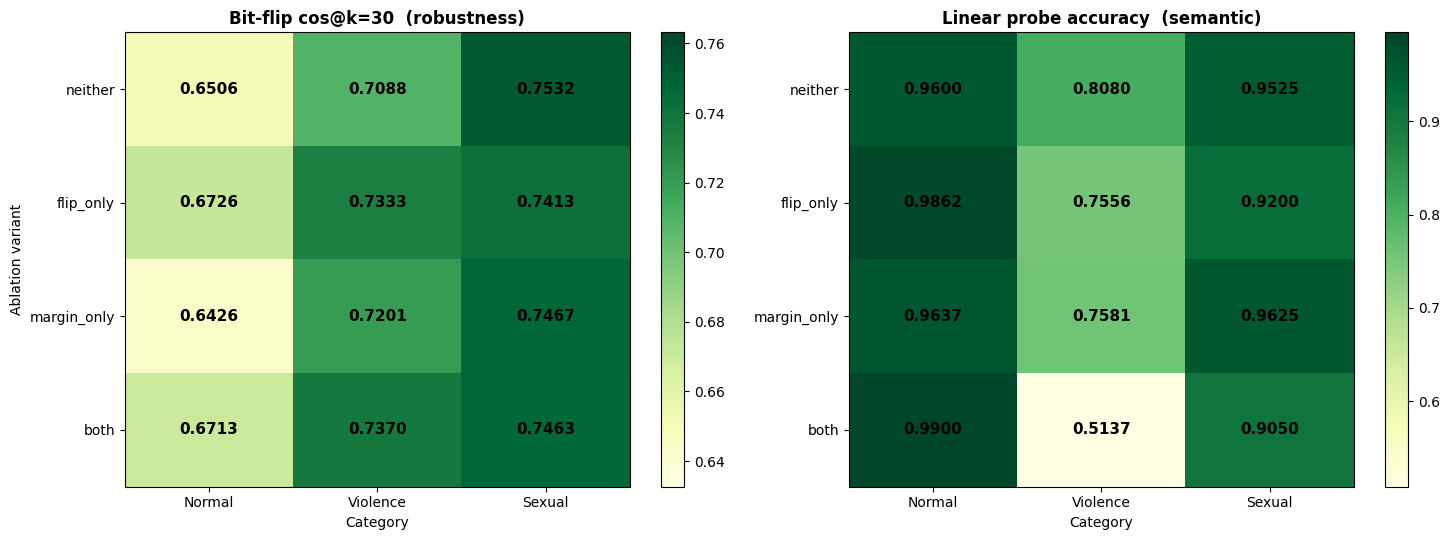}
    \caption{Per-category breakdown of the four ablation variants.
    flip-noise injection improves bit-flip robustness uniformly across
    Normal, Violence, and Sexual categories.}
    \label{fig:ablation_per_cat}
\end{figure}

\Cref{fig:ablation_curves} shows that flip-noise alone is sufficient
to recover most of the bit-flip robustness gain, while
\cref{fig:ablation_per_cat} confirms that this improvement holds
consistently across all three semantic categories.

\section{Hyperparameter Sensitivity (\textit{$\lambda_\text{flip}$, margin, $k_\text{max}$})}
\label{sec:appendix_hp_sweep}
\Cref{tab:hp_sweep} reports a $\lambda_\text{flip}$ sweep
(at fixed $\lambda_\text{margin}\!=\!0.01$) and a 4-cell grid over
$\text{margin}\!\in\!\{0.1,0.5\}$ and $k_\text{max}\!\in\!\{5,15\}$.
Bit-flip robustness at $k\!=\!30$ is essentially saturated across
configurations ($\sigma\!=\!0.005$); linear-probe accuracy varies more
($\sigma\!=\!0.020$).
This indicates that the channel-robustness benefit of flip-noise
training is robust to hyperparameter choices, and that practical
tuning trades off a few percentage points of linear-probe accuracy.

\begin{table}[H]
\centering
\small
\setlength{\tabcolsep}{6pt}
\caption{Hyperparameter sensitivity of channel-aware training.
The robustness metric ($\cos\!@\!k\!=\!30$) is stable across all
configurations; linear-probe accuracy shows mild variation.}
\label{tab:hp_sweep}
\begin{tabular}{lcc}
\toprule
Configuration                                    & $\cos\!@\!k\!=\!30$ & LinProbe \\
\midrule
$\lambda_\text{flip}\!=\!0.05$                   & 0.703 & 0.880 \\
$\lambda_\text{flip}\!=\!0.10$                   & 0.703 & 0.828 \\
$\lambda_\text{flip}\!=\!0.20$                   & 0.704 & 0.896 \\
$\lambda_\text{flip}\!=\!0.50$                   & 0.697 & 0.813 \\
\midrule
$\text{margin}\!=\!0.1, k_\text{max}\!=\!5$      & 0.704 & 0.872 \\
$\text{margin}\!=\!0.1, k_\text{max}\!=\!15$     & 0.713 & 0.917 \\
$\text{margin}\!=\!0.5, k_\text{max}\!=\!5$      & 0.708 & 0.868 \\
$\text{margin}\!=\!0.5, k_\text{max}\!=\!15$     & 0.699 & 0.899 \\
\bottomrule
\end{tabular}
\end{table}

\end{document}